\documentclass[letterpaper]{article} 
\usepackage{aaai25}  
\usepackage{times}  
\usepackage{helvet}  
\usepackage{courier}  
\usepackage[hyphens]{url}  
\usepackage{graphicx} 
\usepackage{natbib}  
\usepackage{caption} 
\usepackage{algorithm}
\usepackage{algorithmicx}
\usepackage{algpseudocode}

\usepackage{newfloat}
\usepackage{listings}
\DeclareCaptionStyle{ruled}{labelfont=normalfont,labelsep=colon,strut=off} 
\floatstyle{ruled}
\newfloat{listing}{tb}{lst}{}
\floatname{listing}{Listing}
\usepackage[utf8]{inputenc} 
\usepackage{booktabs}       
\usepackage{amsfonts}       
\usepackage{nicefrac}       
\usepackage{microtype}      
\usepackage{xcolor}         
\usepackage{xspace}
\usepackage{tablefootnote}
\usepackage{mdframed}
\usepackage{placeins}
\usepackage{multirow}
\usepackage{mathtools}
\usepackage{booktabs}
\usepackage[misc]{ifsym}

\usepackage{amsmath,amsfonts,bm}

\def\eqref#1{equation~\ref{#1}}

\def\1{\bm{1}}

\DeclareMathAlphabet{\mathsfit}{\encodingdefault}{\sfdefault}{m}{sl}
\SetMathAlphabet{\mathsfit}{bold}{\encodingdefault}{\sfdefault}{bx}{n}

\newcommand{\ie}{\textit{i}.\textit{e}.\xspace}
\newcommand{\eg}{\textit{e}.\textit{g}.,\xspace}

\usepackage{dsfont} 
\newcommand{\indicator}[1]{\mathds{1}{#1}}

\newcommand{\modelname}{InverseCoder\xspace}
\newcommand{\xname}{Inverse-Instruct\xspace}

\newcommand{\XName}{Inverse-Instruct\xspace}

\usepackage{colortbl}
\usepackage{tabularray}
\UseTblrLibrary{booktabs}
\SetTblrInner[booktabs]{abovesep=0pt, belowsep=0pt, rowsep=0.5pt}
\SetTblrInner[booktabs]{cells = {cmd=\small}}

\NewTableCommand\seprule{\specialrule{\lightrulewidth,gray8}{2.5pt}{2.5pt}}
\NewTableCommand\uniquerule{\specialrule{\lightrulewidth,gray7,dashed}{2.5pt}{2.5pt}}
\definecolor{lightb}{RGB}{235,245,255}

\usepackage{graphicx}

\usepackage{pifont}%

\usepackage[nointegrals]{wasysym}
\usepackage{enumitem} %
\setlist[itemize]{leftmargin=*}

\usepackage{xspace}
\usepackage{xcolor}
\usepackage{mathtools}

\usepackage[capitalise, noabbrev]{cleveref}
\crefformat{section}{\S#2#1#3}
\Crefformat{section}{\S#2#1#3}

\usepackage{fontawesome}

\definecolor{codegreen}{rgb}{0,0.6,0}
\definecolor{codegray}{rgb}{0.5,0.5,0.5}
\definecolor{codepurple}{rgb}{0.58,0,0.82}
\definecolor{backcolour}{rgb}{0.95,0.95,0.92}
\definecolor{codeblue}{rgb}{0,0,0.7}

\makeatletter
\newcommand\codefontsize{\@setfontsize\codefontsize\@viiipt\@ixpt}
\makeatother

\lstdefinestyle{codestyle}{
    backgroundcolor=\color{backcolour},   
    commentstyle=\color{codegreen},
    keywordstyle=\color{codeblue},
    numberstyle=\tiny\color{codegray},
    stringstyle=\color{codepurple},
    basicstyle=\ttfamily\codefontsize,
    breakatwhitespace=false,         
    breaklines=false,                 
    captionpos=b,                    
    keepspaces=false,                 
    showspaces=false,                
    showstringspaces=false,
    showtabs=false,                  
    tabsize=2,
    numbers=none
}

\newcommand\Passat[1]{\mbox{Pass@{#1}}}
\newcommand\humaneval{HumanEval}

\newcommand\evalplus{EvalPlus}
\newcommand\mbpp{MBPP}

\newcommand\dsonek{DS-1000}

\newcommand\multiple{MultiPL-E}

\newcommand\yestoken{\texttt{YES}\xspace}

\newcommand{\selfinstruct}{\textsc{Self-Instruct}\xspace}
\newcommand{\evolinstruct}{\textit{Evol-Instruct}\xspace}
\newcommand{\ossinstruct}{\textsc{OSS-Instruct}\xspace}

\newcommand\codellamaPyAbbrev{CL}

\newcommand\dscoderAbbrev{DS}
\newcommand\magicoder{\mbox{Magicoder}\xspace}

\newcommand\magicoders{\mbox{Magicoder$\mathcal{S}$}\xspace}

\newcommand\wizardcodergpt{WizardCoder-GPT4\xspace}
\newcommand\wizardcodergptc{WizardCoder-GPT4-\codellamaPyAbbrev{}\xspace}

\newcommand\modelnamed{InverseCoder-\dscoderAbbrev{}\xspace}

\newcommand\evolcode{\texttt{evol-codealpaca-v1}\xspace}

\newcommand\codellamapy{CodeLlama-Python}

\newcommand\dscoder{DeepSeek-Coder}
\newcommand\dscoderbase{DeepSeek-Coder-Base}

\newcommand\llm{LLM\xspace}

\newcommand{\UCAS}{University of Chinese Academy of Sciences}
\newcommand{\SKLP}{SKL of Processors, Institute of Computing Technology, CAS}

\newcommand{\BAIDU}{Baidu Inc., Beijing, China}
\newcommand{\AutoDesk}{Autodesk Research}

\title{\modelname: Self-improving Instruction-Tuned Code LLMs with Inverse-Instruct}
\author{
    Yutong~Wu\textsuperscript{\rm 1, \rm 2},
    Di~Huang\textsuperscript{\rm 1},
    Wenxuan~Shi\textsuperscript{\rm 1, \rm 2}, 
    Wei~Wang\textsuperscript{\rm 3},
    Lingzhe~Gao\textsuperscript{\rm 3} \\
    Shihao~Liu\textsuperscript{\rm 3},
    Ziyuan~Nan\textsuperscript{\rm 1, \rm 2},
    Kaizhao~Yuan\textsuperscript{\rm 1, \rm 2},
    Rui~Zhang\textsuperscript{\rm 1},
    Xishan~Zhang\textsuperscript{\rm 1}, \\
    Zidong~Du\textsuperscript{\rm 1},
    Qi~Guo\textsuperscript{\rm 1},
    Yewen~Pu\textsuperscript{\rm 4},
    Dawei~Yin\textsuperscript{\rm 3},
    Xing~Hu\textsuperscript{\rm 1},
    Yunji~Chen\textsuperscript{\rm 1, \rm 2}\thanks{Corresponding author.} \\
}
\affiliations{
    \textsuperscript{1}\SKLP{}  \\
    \textsuperscript{2}\UCAS{}  \\
    \textsuperscript{3}\BAIDU{} \\ 
    \textsuperscript{4}\AutoDesk{} \\
}

\begin{document}

\maketitle

\begin{abstract}

Recent advancements in open-source code large language models (LLMs) have been driven by fine-tuning on the data generated from powerful closed-source LLMs, which are expensive to obtain. This paper explores whether it is possible to use a fine-tuned open-source model to generate additional data to augment its instruction-tuning dataset. We make two observations: (1) A code snippet can serve as the response to different instructions. (2) Instruction-tuned code LLMs perform better at translating code into instructions than the reverse.
Based on these observations, we propose \xname, a data augmentation technique that uses a fine-tuned LLM to generate additional instructions of code responses from its own training dataset. The additional instruction-response pairs are added to the original dataset, and a stronger code LLM can be obtained by fine-tuning on the augmented dataset. We empirically validate \xname on a range of open-source code models (\eg CodeLlama-Python and DeepSeek-Coder) and benchmarks (e.g., HumanEval(+), MBPP(+), DS-1000 and MultiPL-E), showing it consistently improves the base models.

\end{abstract}

\begin{links}
    \link{Code}{https://github.com/wyt2000/InverseCoder}
    \link{Extended version}{https://arxiv.org/abs/2407.05700}
\end{links}
\section{Introduction}

\begin{figure*}[t]
    \centering
    \includegraphics[width=1\linewidth]{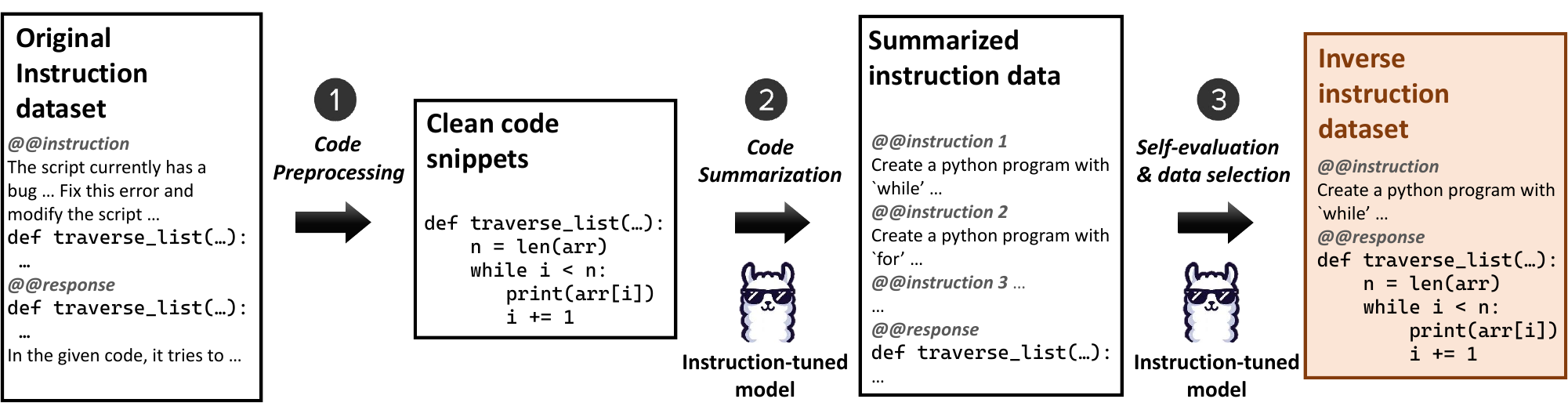}
    \caption{\textbf{The overview of \xname.} 
    \xname utilizes the models' own capability in code summarization to generate an inverse instruction dataset which can further enhance the model's performance.
    \xname consists of three steps, including code preprocessing, code summarization, and self-evaluation \& data selection.}
    \label{fig:overview}
\end{figure*}

Code generation, which aims to generate code that satisfies the user's intent from inputs/outputs or natural language, has been a significant challenge in computer science.
Recently, closed-source LLMs like GPT-3.5 and GPT-4~\cite{gpt4-report} have enabled the generation of general-purpose code (like Python) based on natural language, making them broadly applicable in the fields of programming assistance~\cite{copilot}, computer vision~\cite{suris2023vipergpt,gupta2023visual}, science~\cite{nejjar2023llms}, and embodied intelligence~\cite{liang2023code,ma2023eureka,tang2024worldcoder,wang2023voyager}.

To develop high-performance open-source models, researchers have leveraged these closed-source LLMs to generate datasets of instructions and code, then distilled these datasets into smaller, open-source code LLMs via instruction tuning~\cite{wizardcoder,magicoder,wavecoder,alchemistcoder}.
For example, Code Alpaca~\cite{codealpaca} was fine-tuned on 20K instruction-code pairs generated based on GPT-3.5 with \selfinstruct~\cite{self-instruct}.
\citet{wizardcoder} used \evolinstruct~\cite{wizardlm}, a method that creates a diverse set of instruction data from GPT-3.5 for code generation via evolution heuristics.
\ossinstruct \cite{magicoder} first creates coding problems from the source code snippet, then queries strong LLMs for their corresponding solutions. 
Fine-tuned with 75K GPT-3.5 \ossinstruct data and 110K GPT-4 \evolinstruct data (\ie \evolcode)~\cite{evol-gpt-4}, \magicoders series achieve state-of-the-art (SOTA) results among open-source code models.
These approaches have one thing in common: they heavily rely on generating data by querying stronger closed-source LLMs (\eg GPT-4), which incurs significant additional expenses. Therefore, it is crucial to develop a self-improvement method for open-source models without relying on stronger guidance.

This paper explores how to improve an instruction-tuned code LLM by \emph{querying itself} (rather than querying a closed-source LLM).
We make the following two observations :
\begin{enumerate}
\item A single code snippet can serve as a valid response to multiple instructions.
\item Instruction-tuned code LLMs perform better at translating code into instructions than translating instructions into code (see Section \ref{sec:insights}).
\end{enumerate}
The first observation suggests that an instruction-tuned LLM can generate a new instruction for each response code in its training dataset, thereby expanding the original dataset. The second observation confirms that generating data in this way (Code-to-NL) is more effective than NL-to-Code.

Therefore, we develop \xname, a simple yet effective instruction tuning approach based on self-generating instructions from code snippets (Figure \ref{fig:overview}).
\xname starts with an instruction-code corpus, and a code LLM fine-tuned on it. 
We first clean and extract code snippets from the corpus, then let the code LLM translate these code snippets into new instructions.
Next, we use the code LLM to evaluate and filter consistent instruction-code pairs from the newly generated data.
Finally, the filtered dataset is combined with the original instruction dataset to fine-tune a new model.
The main differences between \xname and previous data generation methods are discussed in Section \ref{sec:other_SFT_methods}.




Using \xname, we develop \modelname, a series of fine-tuned code LLMs that achieve SOTA results. We evaluated \modelname on a wide range of benchmarks (Section \ref{sec:experiments}), including HumanEval(+)~\cite{codex, EvalPlus}, MBPP(+)~\cite{mbpp, EvalPlus}, MultiPL-E~\cite{multiple}, and DS-1000~\cite{ds1000}. Results show that \modelname series surpasses the base models by exploiting the base models' own capability. Specifically, \modelname-DS-6.7B achieves 76.8\% on HumanEval+, 69.0\% on MBPP+, 62.6\% on MultiPL-E, 44.2\% on DS-1000, which are SOTA results across four benchmarks among fully open-source (both model and dataset) models with only 6.7B parameters.

 Our key contributions are introducing \xname, an effective self-improvement instruction tuning approach for code LLMs and presenting a series of code LLMs named \modelname, which achieves SOTA or comparative results on a wide range of benchmarks. 

We organize the structure of the paper as follows:
Section \ref{sec:related_work} introduces related works.
Section \ref{sec:insights} shows the evidence of our observations. 
Section \ref{sec:methods}, \ref{sec:implementation} provide a detailed explanation of our approach (\ie \xname). 
Section \ref{sec:experiments} presents the experiments for our models (\ie \modelname). 
Section \ref{sec:conclusion} concludes with a summary.

\section{Related Work}
\label{sec:related_work}
\subsection{LLMs for Code Generation}

After being pre-trained on a large amount of code, LLMs have demonstrated impressive code generation capabilities. Recently, AI code assistants have become one of the most important applications of LLMs. Technology companies such as OpenAI and Google have developed and released many closed-source large language models, including Codex \cite{codex}, GPT-4 \cite{gpt4-report}, PaLM \cite{PaLM}, and Gemini \cite{gemini}, which have achieved outstanding performance on code generation benchmarks.

In addition to closed-source models, there are also some available open-source models whose weights and training data are available to the public, such as CodeGen \cite{codegen}, CodeGeeX \cite{CodeGeeX}, AlphaCode \cite{AlphaCode}, CodeT5 series \cite{codet5}, StarCoder series \cite{StarCoder, starcoder2}, CodeLlama \cite{codellama}, DeepSeek-Coder \cite{DeepSeekCoder} and  CodeQwen \cite{CodeQwen}. These open-source code models have shown notable advancements in code-related tasks, but there is still a gap compared to the most advanced code LLMs. 


\subsection{Instruction-Tuned Code LLMs}
\label{sec:other_SFT_methods}
Instruction tuning is a method for further enhancing the instruction-following capability of pre-trained LLMs. It has been widely applied to the LLMs for general tasks including
T5 \cite{t5} and FLAN \cite{FLAN}. 

For code LLMs, OctoPack \cite{OctoPack} and PIE \cite{pie} extracted high-quality data from human-written instructions and code. Fine-tuning with these data has significantly enhanced the program generation capabilities of the base models.

However, obtaining high-quality human-written instruction datasets is usually laborious. Researchers have attempted to employ stronger closed-source LLMs to generate both new instructions and responses for instruction-tuning. Specifically, CodeAlpaca \cite{codealpaca} sampled tasks from a seed task pool and prompted a stronger LLM to generate instruction-tuning data based on the seed tasks. WizardCoder \cite{wizardcoder} prompted a stronger LLM to generate more complex instructions and the corresponding responses. Magicoder \cite{magicoder} used a stronger LLM to create problems and code solutions based on open-source code snippets, as the seed code snippets offer controllability and diversity to the generation. WaveCoder \cite{wavecoder} used a stronger LLM to both generate and discriminate the instruction-response pair for different coding tasks (\eg code summarization and code repair). AlchemistCoder \cite{alchemistcoder} employed a stronger LLM to add more details for existing instructions.  

The main differences between our method and the aforementioned related works are:

\begin{itemize}
    \item We focus on the \textbf{self-improvement} of open-source code models rather than relying on stronger guidance (such as human annotation or advanced LLMs like GPT-4).
    \item We generate new data by \textbf{converting code to instructions} from existing datasets rather than generating code from instructions.
\end{itemize}




\section{Sanity Check: Code-to-NL vs. NL-to-Code}
\label{sec:insights}

In this section, we validate our observation that instruction-tuned code LLMs perform better at translating code into instructions (\ie, Code-to-NL) than translating instructions into code (\ie, NL-to-Code) through an experiment.

\begin{table}[t]
    \centering
    \setlength{\tabcolsep}{0.5mm}
    \begin{tabular}{lcc}
    \toprule
    Generation Method  & WC-CL-7B & WC-DS-6.7B \\
    \midrule
    NL $\rightarrow$ Code  &  62.4         & 70.2         \\
    Code $\rightarrow$ NL $\xrightarrow{\text{GPT-4}}$ Code  & \textbf{74.3} & \textbf{79.0}         \\
    Code $\rightarrow$ NL $\xrightarrow{\text{Humans}}$ Code & \textbf{86.7}         & \textbf{80.0} \\
    \bottomrule
    \end{tabular}
     \caption{\textbf{\Passat{1} (\%) results on MBPP+ in observation checking experiments.} The abbreviations ``WC-CL-7B'' and ``WC-DS-6.7B'' refer to the instruction-tuned models WizardCoder-GPT4-CL and WizardCoder-GPT4-DS. Line 1 represents the evaluation of NL-to-Code for instruction-tuned open models. Lines 2 and 3 evaluate Code-to-NL by leveraging GPT-4 and humans to convert NL into its equivalent code, then assess its correctness against the original code. We removed the problems that GPT-4 was unable to give executable code for them.}
    \label{tab:observation_check_exp}
\end{table}

We first select a manually written set of correctly matched NL-Code pairs $\{x, y\}$ with unit tests and prompted a fine-tuned code LLM to convert $x$ into new code $y'$ and $y$ into new NL $x'$ separately.
Then, We use the following metrics to quantify the model's performance in the two tasks:

\begin{itemize}
    \item For NL-to-Code, we use unit tests to evaluate the functional correctness of generated code $y'$ against original code $y$.
    \item For Code-to-NL, we convert generated NL $x'$ to an equivalent code snippet $\hat{y}$ by humans and a stronger code LLM. Then we measured the functional correctness of $\hat{y}$ by unit tests.
\end{itemize}

Specifically, we use the problem-answer pairs with unit tests in a basic Python generation benchmark MBPP+ \cite{EvalPlus} as matched NL-Code pairs $\{x, y\}$. For NL-to-Code, we took all 378 problems in the benchmark for evaluation. For Code-to-NL, we first select 30 problems for humans to write the equivalent code of the generated NL, and then we employ GPT-4 to finish this task for all problems.




We evaluate two instruction fine-tuned code LLMs (\ie, WizardCoder-GPT4-CL and WizardCoder-GPT4-DS, which are instruction-tuned by 110K GPT-4 dataset \evolcode). The results are shown in Table \ref{tab:observation_check_exp}. From the table, we conclude that (Code $\rightarrow$ NL) is better than (NL $\rightarrow$ Code), showing that code LLMs perform better in code summarization than in code generation.
\section{\XName: Data Augmentation via Code Summarization}
\label{sec:methods}

In this section, we will introduce \xname, a data augmentation method that can obtain more instruction data through the model's own capabilities.
The overall illustration of \xname is shown in Figure \ref{fig:overview}. 
\xname is founded on the following two observations:
(1) The same code can be considered as a response to different instructions, which expands the dataset effectively.
(2) Converting formal language (i.e., code) into informal language (i.e., natural language) is generally more straightforward than the reverse.

The whole data generation process contains three stages: (1) Code preprocessing. (2) Code summarization. (3) Self-evaluation and data selection.
In code preprocessing, we preprocess the code data by filtering clean code snippets $\{y_i^*\}$ from an off-the-shelf instruction tuning dataset $\{(x_i, y_i)\}$ (\eg \evolcode).
Subsequently, in code summarization, we prompt an instruction fine-tuned code LLM $M$ (\eg \wizardcodergptc) to convert the clean code snippets $\{y_i^*\}$ to multiple new instructions $\{x_{ij}^*\}$.
Then, in self-evaluation and data selection, we use the same code LLM $M$ to select the best instruction $x_{i}^{**}$ in $\{x_{ij}^*\}$. The selected instructions $\{x_{i}^{**}\}$ are combined with the original code snippets $\{y_i^*\}$ to construct a new instruction tuning dataset $\{(x_{i}^{**}, y_i^*)\}$.
Finally, we fine-tune the base code LLM with the instruction data $\{(x_{i}^{**}, y_i^*)\} \cup \{(x_i, y_i)\}$ to obtain a stronger code LLM (\ie \modelname).
Details of the three steps are illustrated below.

\subsection{Code Preprocessing}

The first step is to preprocess the existing code data and get clean code snippets $\{y_i^*\}$.
This is because the Code-to-NL capabilities of code LLMs can only be fully utilized with clean code, whereas the response data $\{y_i\}$ in the original dataset typically contains a lot of noise, such as natural language responses.

We select data with code snippet $\{y_i^*\}$ from the original $\{y_i\}$ with the following two steps:
\begin{enumerate}
\item \textbf{Filtering responses}. We first collect responses that contain the marker of the code block (\ie \`{}\`{}\`{}), which indicates that there are code snippets in the response. The remaining data might contain clean code without any code markers, so then we collect the responses that can pass syntax checking.
\item \textbf{Extracting code}. After filtering responses with code snippets, we remove the natural language surrounding the code snippets to make it easier for the model to summarize. If there are multiple parts of code in the original response, we only keep the first part, since the following parts are usually test cases or using examples. 
\end{enumerate}

At the end of code preprocessing, we obtain clean code snippets $\{y_i^*\}$ for summarization.

\subsection{Code Summarization}
After filtering, we employ the code LLM $M$ to generate a certain number of corresponding instructions $\{x_{ij}^*\}$ for each code snippet in $\{y_i^*\}$ by summarizing its functionality. During the summarization process, we randomly choose different instruction prefixes for the prompt to enhance the diversity of the instructions. 

In this way, we have obtained new pairs of natural language and code $\{(x_{ij}^*, y_i^*)\}$.

\subsection{Self-evaluation and Data Selection}
We noticed that code LLM $M$ might make mistakes during the code summarization process.
Therefore, we utilize $M$ itself to evaluate $\{(x_{ij}^*, y_i^*)\}$ and select the most appropriate instruction.


Data selection is typically performed by powerful LLMs such as GPT-4 because these models possess excellent instruction-following capabilities, enabling them to understand complex filtering rules~\cite{Wang2024ASO}. However, the instruction-following capabilities of code LLMs are often weaker, making it difficult to conduct effective selection. (See the comparison experiments in Section \ref{app:compare_with_other_selection_methods}).

Inspired by AutoMathText \cite{automathtext},
we use the pseudo-probability of \yestoken token given by the code LLM $M$ as an indicator of the instruction quality rather than a score in textual format. Specifically, we concatenate the generated instructions $\{x_{ij}^*\}$ and the original code snippets $\{y_i^*\}$ as problem-answer pairs $\{(x_{ij}^*, y_i^*)\}$ . 
Then, we ask $M$ to evaluate the correctness of each answer under the given problem and calculate the pseudo-probability of \yestoken using the logits of the first token given by $M$.
The formula for calculating the pseudo-probability is shown as follows \cite{automathtext}:

\begin{equation*}
    \operatorname{LM-Score}(\cdot) = \frac{\operatorname{exp}(\operatorname{logit}(\text{`YES'}))}{\operatorname{exp}(\operatorname{logit}(\text{`YES'})) + \operatorname{exp}(\operatorname{logit}(\text{`NO'}))}
\end{equation*}

After evaluation, we select the instruction with the highest score $x_{i}^{**}$ for each response in $\{y_i^*\}$ to obtain a new training dataset $\{(x_{i}^{**}, y_i^*)\}$.

\section{Implementation Details}
\label{sec:implementation}

\paragraph{The original instruction tuning dataset.}
In this work, we mainly use \evolcode as our original instruction tuning dataset $\{(x_i, y_i)\}$, which is widely used for instruction tuning of code LLMs~\cite{magicoder,wavecoder,alchemistcoder}. 
It contains 111183 instruction-response pairs generated by \evolinstruct using GPT-4. 
Following Magicoder \cite{magicoder}, \evolcode is decontaminated by removing data that contain docstrings or solutions from HumanEval~\cite{codex}, MBPP~\cite{mbpp}, MultiPL-E~\cite{multiple}, and DS-1000~\cite{ds1000}, which are used to evaluate \modelname. We apply the same decontamination method to the newly generated instruction data $\{(x_i^{**}, y_i^*)\}$.

\paragraph{Training for original Code LLM.}
We take \codellamapy-13B, \codellamapy-7B \cite{codellama} and \dscoderbase-6.7B \cite{DeepSeekCoder} as base models. To obtain the beginning code LLM $M$ (hereinafter called \wizardcodergpt), we fine-tune the base models on \evolcode for $2$ epochs using $8$ NVIDIA A100-40GB SMX GPUs. We set the initial learning rate at $5e-5$ with $15$ warmup steps and a linear learning rate scheduler. We use Adafactor \cite{Adafactor} as our optimizer and choose a batch size of $512$ with a sequence truncation length of $1024$.

\paragraph{Instruction data collection.}
We use the vLLM inference framework \cite{vllm} for code summarization and instruction selection on the same GPUs as training. We generate $10$ instructions $\{x_{ij}^*\}_{j=1}^{10}$ for each code snippet in the code summarization stage. For each instruction-response pair, the self-evaluation and data selection process is conducted by prompting the beginning code LLM $M$ with greedy decoding. We choose the instruction with the highest pseudo-probability of \yestoken as the best-generated instruction for each response.

\paragraph{Training for \modelname.}

Following \magicoders \cite{magicoder}, we first fine-tune the base models on the new dataset $\{(x_i^{**}, y_i^*)\}$ with 90363 instruction-response pairs (generated by the original Code LLM $M$) for $1$ epoch, then we continue to fine-tune the models with the original dataset $\{(x_i, y_i)\}$ (generated by GPT-4) for $2$ epochs to obtain \modelname. 
The hyperparameters are the same as the training process for the original code LLM $M$. The instruction tuning prompt is aligned with \magicoders. 

\section{Experiments}
\label{sec:experiments}
We conduct a series of experiments to investigate these topics:

\begin{enumerate}

    \item \modelname's performance on benchmarks (Sec. \ref{sec:main_results}).
    
    \item Impact of each stage in \xname (Sec. \ref{sec:ablation_study}).

    \item Impact of dataset size scaling (Sec. \ref{app:data_scaling}).

    \item Is \xname effective on other datasets (Sec. \ref{sec:different_original_dataset})?

    \item Comparison with other data selection methods (Sec. \ref{app:compare_with_other_selection_methods}).

    \item Does selecting multiple self-generated instructions for each response lead to further improvement (Sec. \ref{sec:select_multiple_instructions})?

    \item Can \xname be repeatedly applied to \modelname to achieve multi-round optimization (Sec. \ref{sec:multi_round})?

    \item Can \xname be further optimized by using additional self-generated code as responses (Sec. \ref{sec:code_evol})?




\end{enumerate}




\subsection{Main Results}
\label{sec:main_results}

We train \modelname on three base models with different parameter sizes and evaluate them on four benchmarks widely used for code LLMs, including Python text-to-code generation, multilingual coding, and data-science code generation. The results show that \textbf{the performance of SOTA code LLMs can continue to improve by \xname}. 

\subsubsection{Baselines.}
We compare the performance of our models with a wide range of baselines including:
\begin{enumerate}

\begin{table}[t]
\centering
\setlength{\tabcolsep}{1mm}
\begin{tabular}{l|c|l}
\toprule
Model               & Common Data                 & Specific Data            \\
\midrule
WizardCoder-GPT-4   & \multirow{5}{*}{110K GPT-4} & 0K (baseline)               \\
\magicoders          &                             & 75K GPT-3.5                 \\
WaveCoder-Ultra     &                             & 20K GPT-4                   \\
AlchemistCoder &                              & $>$ 80K GPT-3.5                 \\
 \rowcolor[rgb]{0.925,0.925,0.925}InverseCoder (ours) &                              & 90K \textbf{self-generated} \\
\bottomrule
\end{tabular}
\caption{\textbf{Training data size of different instruction-tuned code LLMs.} It is worth noting that only \modelname is trained by self-generated data, which is easier to obtain at a lower cost.}
\label{tab:data_scale_with_other_models}
\end{table}

\begin{table}[!t]
\setlength{\tabcolsep}{0.5mm}
\centering
\begin{tabular}{lccc}
\toprule
Model                    & HumanEval~(+) & MBPP~(+)     \\
\midrule
\multicolumn{3}{l}{\it (Closed-source Models)}                   \\
GPT-4-Turbo (April 2024) & \textbf{90.2 (86.6)}  & \textbf{85.7 (73.3)} \\
GPT-3.5-Turbo (Nov 2023) & \underline{76.8} (\underline{70.7})  & \underline{82.5} (\underline{69.7}) \\

\midrule 
\multicolumn{3}{l}{\it (Based on \codellamapy-13B)}                   \\
\codellamapy-13B           & 42.7 (38.4)  & 63.5 (52.6) \\
WizardCoder-GPT4-CL-13B      & \underline{76.8} (\underline{70.7})  & \underline{73.5} (\underline{62.2}) \\
\rowcolor[rgb]{0.925,0.925,0.925}InverseCoder-CL-13B (ours)   & \textbf{79.9 (74.4)}  & \textbf{74.6 (63.0)} \\

\midrule 
\multicolumn{3}{l}{\it (Based on \codellamapy-7B)}                   \\
\codellamapy-7B          & 37.8 (35.4)  & 59.5 (46.8) \\
\magicoders-CL-7B           & 70.7 (67.7)  & \textbf{70.6 (60.1)} \\
AlchemistCoder-CL-7B        & 74.4 (68.3) & 68.5 (55.1) \\      
WizardCoder-GPT4-CL-7B      & \underline{72.6} (\underline{68.9})  & \underline{69.3} (\underline{59.3}) \\
\rowcolor[rgb]{0.925,0.925,0.925}InverseCoder-CL-7B (ours)   & \textbf{76.2 (72.0)}  & \textbf{70.6 (60.1)} \\

\midrule
\multicolumn{3}{l}{\it (Based on DeepSeek-Coder-6.7B)}                   \\
DeepSeek-Coder-6.7B      & 47.6 (39.6)  & 72.0 (58.7) \\
\magicoders-DS-6.7B           & 76.8 (71.3)  & \textbf{79.4} (\textbf{69.0}) \\
WaveCoder-Ultra-DS-6.7B       & 75.0 (69.5)  & 74.9 (63.5) \\
AlchemistCoder-DS-6.7B        & \textbf{79.9} (75.6) & 77.0 (60.2) \\
WizardCoder-GPT4-DS-6.7B      & \underline{77.4} (\underline{73.2})  & 77.8 (\underline{67.5}) \\
\rowcolor[rgb]{0.925,0.925,0.925}InverseCoder-DS-6.7B (ours)   & \textbf{79.9 (76.8)}  & \underline{78.6} (\textbf{69.0}) \\

\bottomrule
\end{tabular}
 \caption{\textbf{\Passat{1} (\%) results of different \llm{s} on \humaneval{}~(+) and \mbpp{}~(+) computed with greedy decoding.} The abbreviations ``CL'' and ``DS'' refer to the base models \codellamapy~and \dscoder, respectively.
 We report other results consistently from the \evalplus~\cite{EvalPlus} Leaderboard in August 2024 and Magicoder \cite{magicoder} paper. 
 }
\label{tab:python-text2code}

\end{table}
\begin{table}[t]
\centering
\setlength{\tabcolsep}{1.0mm}
\fontsize{9}{11}\selectfont
\begin{tabular}{lccccccc}
\toprule
Model                   & Java & JS & C++  & PHP  & Swift & Rust & Avg. \\

\midrule
\multicolumn{8}{l}{\it (Based on CodeLlama-Python-13B)} \\
WizardCoder-GPT4*    & \textbf{55.4}& \underline{64.2} & \underline{55.9} & \underline{52.0} & \underline{49.9} & \underline{53.4} & \underline{55.1} \\
\rowcolor[rgb]{0.925,0.925,0.925} InverseCoder (ours)* & 54.5 & \textbf{65.4} & \textbf{58.1} & \textbf{55.3} & \textbf{52.5} & \textbf{55.6} & \textbf{56.9} \\

\midrule
\multicolumn{8}{l}{\it (Based on CodeLlama-Python-7B)} \\
CodeLlama-Python        & 29.1 & 35.7       & 30.2 & 29.0 & 27.1  & 27.0 & 29.7 \\
\magicoders*         & \underline{49.8} & \textbf{62.6} & 50.2 & \underline{53.3} & 44.9 & 43.8 & 50.8 \\
WizardCoder-GPT4*    & \textbf{50.4} & 60.7 & \underline{50.6} & 51.6 & \underline{45.6} & \textbf{48.2} & \underline{51.2} \\
\rowcolor[rgb]{0.925,0.925,0.925}InverseCoder (ours)* & 48.7 & \underline{61.9} & \textbf{52.6} & \textbf{55.2} & \textbf{53.0} & \underline{46.1} & \textbf{52.9} \\

\midrule
\multicolumn{8}{l}{\it (Based on DeepSeek-Coder-6.7B)} \\
\magicoders* & 59.6 & \underline{69.8} & \underline{70.0} & \textbf{64.4} & \textbf{54.4} & 53.6 & \underline{62.0} \\
WizardCoder-GPT4*    & \textbf{61.4} & 66.4 & 68.7 & 61.8 & 52.6 & \underline{56.1} & 61.2 \\
\rowcolor[rgb]{0.925,0.925,0.925}InverseCoder (ours)*  & \underline{60.7} & \textbf{70.1} & \textbf{70.5} & \underline{63.6} & \underline{53.0} & \textbf{57.4} & \textbf{62.6} \\

\bottomrule
\end{tabular}
\caption{\textbf{\Passat{1} (\%) results of different \llm{s} on \multiple.} The models marked with (*) are evaluated with the same prompt format as training and the same hyperparameter as Magicoder. We report other results consistently from Magicoder paper.}
\label{tab:multilang}
\end{table}

\begin{table}[t]
\centering
\setlength{\tabcolsep}{0.45mm}
\fontsize{9}{11}\selectfont
\newcommand\rotateHeader[1]{\rotatebox{0}{#1}}
  \begin{tabular}{lcccccccc}
    \toprule
    Model                  & plt. & np. & pd. & torch & scipy & sklearn & tf. & All \\

    \midrule
    \multicolumn{9}{l}{\it (Based on \codellamapy-13B)} \\
    WizardCoder-GPT4  & \textbf{56.1} & \underline{52.2} & \underline{30.3} & \underline{43.0} & \textbf{25.2} & \underline{49.5} & \underline{40.0} & \underline{42.1} \\
     \rowcolor[rgb]{0.925,0.925,0.925}InverseCoder (ours) & \underline{53.0} & \textbf{54.3} & \textbf{32.1} & \textbf{50.9} & \underline{22.5} & \textbf{50.5} & \textbf{43.8} & \textbf{43.1} \\

    \midrule
    \multicolumn{9}{l}{\it (Based on \codellamapy-7B)} \\
    CodeLlama-Python     & \underline{55.3} & 34.5  & 16.4   & 19.9    & 22.3  & 17.6    & 28.5       & 28.0    \\
    WizardCoder       & 53.5       & 34.4  & 15.2   & 25.7    & 21.0  & 24.5    & 28.9       & 28.4    \\
    \magicoders          & \textbf{55.9} & 40.6  & \underline{28.4}   & \underline{40.4}    & 28.8  & \underline{35.8}   & 37.6      & 37.5    \\
    WizardCoder-GPT4     & 51.5       & \underline{46.9}  & \textbf{29.9}   & \textbf{43.6}    & \textbf{34.9}  & \textbf{41.9}    & \underline{39.0}       & \textbf{40.2}    \\
     \rowcolor[rgb]{0.925,0.925,0.925}InverseCoder (ours)  & 54.2  & \textbf{48.6}  & 27.4   & 38.0    & \underline{34.0}  & \textbf{41.9}    & \textbf{40.3}       & \underline{39.9}    \\
    
    \midrule
    \multicolumn{9}{l}{\it (Based on DeepSeek-Coder-6.7B)} \\
    \magicoders & \underline{54.8} & \underline{48.9} & \underline{30.0}	& 49.2 & 27.3 & 44.7 & \underline{41.2} &	41.2 \\
    WizardCoder-GPT4  & 53.8       & \textbf{53.9}  & 28.0   & \underline{49.3}    & \textbf{30.4}  & \underline{45.7}    & \textbf{44.4}       & \underline{42.2}    \\
 \rowcolor[rgb]{0.925,0.925,0.925}    InverseCoder (ours) & \textbf{55.5}       & \textbf{53.9}  & \textbf{32.3}   & \textbf{56.7}    & \underline{30.0}  & \textbf{50.3}    & 33.9       & \textbf{44.2}    \\

    \bottomrule
  \end{tabular}
\caption{\textbf{\Passat{1} (\%) results on \dsonek{}} including seven data science libraries: Matplotlib (plt.), Numpy (np.), Pandas (pd.), Pytorch, Scipy, Sklearn and Tensorflow (tf.). We evaluate our models in the same prompt and hyperparameters as \magicoder. We report other results from \magicoder paper.
}
\label{tab:ds1000}
\end{table}

\item \textbf{Base Models}: Three base models mentioned in Section \ref{sec:implementation}. We compare \modelname with them to show the absolute improvement of the whole instruction-tuning process.

\item \textbf{WizardCoder-GPT4}: The beginning code LLMs in our data generation process, which are only trained by the original instruction-tuning dataset (\ie, \evolcode). We compared \modelname with them to show the improvement brought by \xname.  

\item \textbf{Other Open Source Instruction-Tuned Code LLMs}: Instruction-tuned code models in related works, including \magicoders \cite{magicoder}, WaveCoder-Ultra-DS \cite{wavecoder} and AlchemistCoder \cite{alchemistcoder}. They are trained on additional data generated by stronger closed-source LLMs (\eg GPT-3.5) in addition to \evolcode.  

The comparison of training data size is shown in Table \ref{tab:data_scale_with_other_models}. The actual data consumption of \modelname should be mainly measured by the scale of the \textbf{original training dataset (110K)} since the cost of self-generating data is much lower than generating data by querying closed-source LLMs \cite{ScalingDownToScaleUp}.

\item \textbf{Closed-source LLMs}: GPT-3.5 \cite{chatgpt} and GPT-4 \cite{gpt4-report} to show the gap between \modelname with the advanced closed-source LLMs.

\end{enumerate}

\subsubsection{\xname improves general Python code generation capabilities.}

We use HumanEval(+) and MBPP(+) \cite{EvalPlus}, the enhanced versions of two Python code generation benchmarks \cite{codex,mbpp}, to evaluate the text-to-code capability of \modelname. Each benchmark provides a set of tasks with natural language descriptions as prompts for the code LLM to generate function-level code, which is then validated using pre-prepared test cases.

We use the pass@1~\cite{codex} score to compare the code generation capability among different models. The results are shown in Table~\ref{tab:python-text2code}, which demonstrate that \modelname makes a significant improvement over \wizardcodergpt in Python code generation capability. Furthermore, \modelnamed-6.7B has an outstanding performance in HumanEval/HumanEval+, which surpasses all open-source models with a similar scale of weights.

\subsubsection{The improvement of \xname is reflected across multiple programming languages.}

Besides Python, we evaluate the code generation capabilities of other six mainstream programming languages for \modelname on MultiPL-E benchmark \cite{multiple}. We generate and evaluate code of different programming languages under the inference prompt format aligned with the prompt we used in the training process.

Table \ref{tab:multilang} shows the performances of \modelname and other models on MultiPL-E. The results reveal that the capabilities of \modelname to generate code in multiple mainstream programming languages are improved over \wizardcodergpt. 

\subsubsection{\xname also leads to enhancement in data science code generation tasks.}

To show the capability of \modelname for complex programming problems in realistic applications, we evaluate it on DS-1000 benchmark \cite{ds1000}, which comprises $1000$ different data science workflows across seven libraries. Following \citet{magicoder}, we evaluate our model only on the completion mode.

The results in Table \ref{tab:ds1000} show that the average performances of \modelname-CL-13B and \modelname-DS-6.7B in the data science code generation tasks are enhanced, which implies that \xname can help to improve the code generation capability of the original model in realistic tasks beyond basic programming problems.

\subsection{Ablation Study}
\label{sec:ablation_study}
\begin{table}[t]
\centering
\setlength{\tabcolsep}{1mm}
\begin{tabular}{lcc}
\toprule
Method                                 & HumanEval(+) & MBPP(+)     \\
\midrule
Gen. + Eval.  & 70.7 (67.1)  & \textbf{70.9} (\textbf{60.1}) \\
Pre.                             & 72.6 (68.9)  & 69.8 (\underline{59.8}) \\
Pre. + Sum.                       & \underline{75.6} (\underline{71.3})  & 68.0 (58.2) \\
Pre. + Sum. + Eval. (ours)                 & \textbf{76.2} (\textbf{72.0})  & \underline{70.6} (\textbf{60.1}) \\
\bottomrule
\end{tabular}
\caption{\textbf{\Passat{1} (\%) results on HumanEval+ and MBPP+ in ablation studies.} Preprocessing (Pre.), Summarization (Sum.) and Evaluation (Eval.) correspond to the three steps in our method. Generation (Gen.) represents regenerate responses for each instruction.}
 \label{tab:ablation}
\end{table}
\begin{figure}[t]
  \centering
\includegraphics[width=1\linewidth]{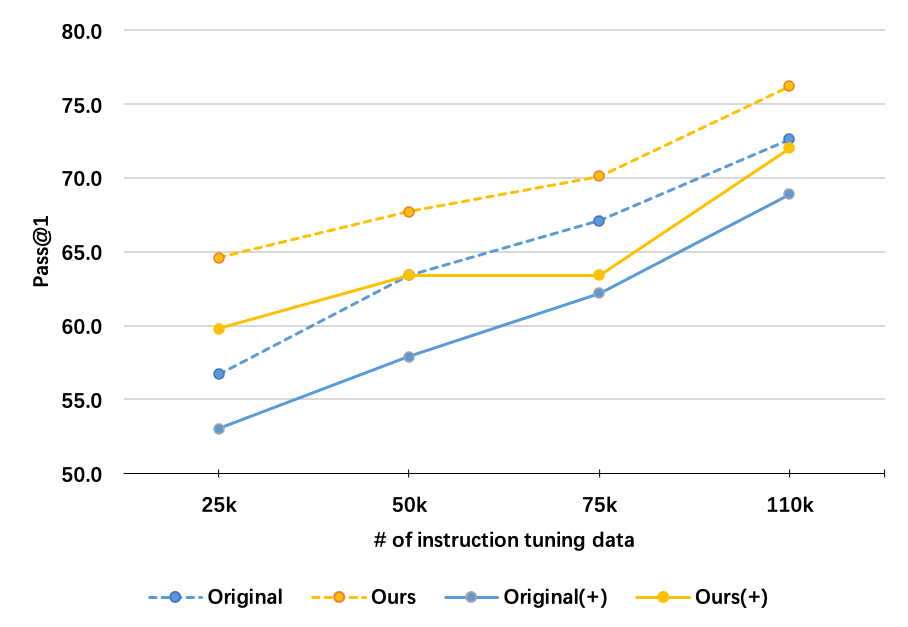}
\caption{\textbf{Impact of data scaling.} The dashed line represents HumanEval and the solid line represents HumanEval+. Legend ``Original'' and ``Ours'' represent the original models and the models improved by \xname. }
\label{fig:scaling}
\end{figure}
\begin{table}[t]
    \centering

        \begin{tabular}{lcc}
        \toprule
        Model           & HumanEval (+) & MBPP (+)    \\
        \midrule
        Magicoder-DS    & \underline{66.5} (\underline{60.4})   & \underline{75.4} (\underline{61.9}) \\
\rowcolor[rgb]{0.925,0.925,0.925}    InverseCoder-DS-OSS & \textbf{69.5 (64.0)}   & \textbf{77.0 (66.1)} \\
        \bottomrule
        \end{tabular}
    
    \caption{\textbf{Performance improvement of \xname when applied to Magicoder-OSS-Instruct-75K.}}
    \label{tab:oss_exp}
\end{table}

We conduct a series of ablation experiments to analyze the utility of code summarization and data selection steps in our method. We use \codellamapy-7B~as the base model in the following experiments with the same training settings as \modelname and present the results in Table \ref{tab:ablation}. The ablation experiments are in three aspects: 

\paragraph{\xname outperforms the NL-to-Code data generation method (Gen. + Eval.).}
We regenerate 10 responses $\{y_{ij}\}_{j=1}^{10}$ for each instruction $x_i$ in the original training dataset and apply the same self-evaluation method to select the best responses. It shows that the code summarization step provides overall better performance than generating responses from instructions.

\paragraph{Performance improvement comes not only from the preprocessing step (Pre.).}
We only apply preprocessing to the responses in the original dataset $\{(x_i, y_i)\}$ to obtain a cleaned dataset $\{(x_i, y_i^*)\}$. We train the models with the cleaned dataset and the original one to show the improvement from preprocessing is minor.

\paragraph{The self-evaluation and data selection step also plays a role in \xname (Pre. + Sum.).}
To study the role of self-evaluation and data selection, we generate only one instruction for each response in the code summarization step without any selection. The results show that self-evaluation and selection are also helpful to performance improvement.

\subsection{Data Scaling}
\label{app:data_scaling}

\paragraph{\xname is effective across different data scales.}
We conduct a series of experiments to explore the data scaling law of \xname.
Specifically, we randomly select 25K, 50K, and 75K instruction-response pairs from the original dataset and train 3 weaker original models with them. 
Then, we apply \xname for the original models.
It is shown that the performances of the models are all improved by \xname at different scales of data (Figure \ref{fig:scaling}).

\subsection{Impact of Original Dataset}
\label{sec:different_original_dataset}
\paragraph{\xname is effective across different original datasets.} We apply \xname to Magicoder-OSS-Instruct-75K \cite{magicoder}, a smaller dataset generated by GPT-3.5.
The results (Table \ref{tab:oss_exp}) show that performance is still improved even with a smaller and lower-quality original dataset, demonstrating the robustness of \xname.

\subsection{Alternative Data Selection Methods}
\label{app:compare_with_other_selection_methods}

\begin{table}[t]
\centering
\begin{tabular}{ll}
\toprule
Data-Selection Method        & HumanEval (+)                     \\
\midrule
Random Selection 	          & 72.6 (68.3) \\
Textual Score                 & \underline{73.8} (\underline{69.5})                       \\
Lowest Perplexity             & 70.1 (67.7)                       \\
Highest Perplexity            & 70.7 (67.7)                       \\
 \rowcolor[rgb]{0.925,0.925,0.925}YES Pseudo-probability (ours) & \textbf{76.2 (72.0)} \\              
\bottomrule
\end{tabular}
\caption{\textbf{Comparison of our data selection method with alternatives (for CL-7B).}}
\label{tab:comparsion_other_selection_methods}
\end{table}

\begin{table}[t]
    \centering

        \begin{tabular}{ccc}
        \toprule
        Selected Instructions & HumanEval (+) & MBPP (+)     \\
        \midrule
\rowcolor[rgb]{0.925,0.925,0.925} Top-1 (ours)                & \textbf{76.2 (72.0)}   & \textbf{70.6 (60.1)}  \\
        Top-3                       & \underline{70.1} (\underline{67.1})	& \underline{68.0} (\underline{58.5})  \\
        Top-5                       & \underline{70.1} (65.2)	& 61.9 (53.4)  \\
        \bottomrule
        \end{tabular}
    
    \caption{\textbf{Performance comparison of the models (CL-7B) trained with different numbers of selected instructions.} ``Top-k'' means that for each response, we select the instructions with the top k highest pseudo-probability.}
    \label{tab:top-k}
\end{table}

\paragraph{Our data selection method outperforms alternatives.} We compare our data selection method which is based on the pseudo-probability of \yestoken with the three alternatives:
\begin{enumerate}
\item Randomly selecting one instruction from all synthetic candidates corresponding to each response.
\item Using textual format scores (1-5) provided by the LLM itself as an indicator. If no textual score is given, assign a default score of 3.
\item Using the sentence perplexity of the response code under different instructions as an indicator. We select the data with the highest and lowest perplexity respectively.
\end{enumerate}
The results are shown in Table \ref{tab:comparsion_other_selection_methods}, demonstrating the pseudo-probability method's efficiency.

\subsection{Selecting Multiple Self-Generated Instructions}
\label{sec:select_multiple_instructions}
 \paragraph{Selecting multiple self-generated instructions for a single response will harm the model's performance.} We select the top-k scoring instructions for each response. The results in Table \ref{tab:top-k} indicate that the model's performance declines as the number of selected instructions increases. This suggests that open-source code LLMs are not capable of generating a large number of correct instructions, which is why we only select the best instructions in our method.

\subsection{Multi-Round Optimization for \modelname}
\label{sec:multi_round}

\paragraph{Repeatedly applying \xname to \modelname does not significantly improve performance.} 
We replace the original model with \modelname in the pipeline of \xname and train a new model with the data generated by \modelname. The performance results (Table \ref{tab:multi-round}) show no significant improvement, which confirms the phenomenon of model collapse caused by repeatedly training on self-generated data \cite{AIModelCollapse}.

\begin{table}[t]
    \centering

        \begin{tabular}{lcc}
        \toprule
        Model & HumanEval (+) & MBPP (+)     \\
        \midrule
        InverseCoder-CL-7B &	\textbf{76.2} (\textbf{72.0}) &	\textbf{70.6} (\underline{60.1}) \\
        InverseCoder-CL-7B-V2 & \underline{75.0} (\underline{70.1}) &	\textbf{70.6} (\textbf{60.6}) \\
        \bottomrule
        \end{tabular}
    
    \caption{\textbf{Performance diffenernce when applying \xname to \modelname again.} ``V2'' means models trained with the data generated by \modelname.}
    \label{tab:multi-round}
\end{table}

\subsection{Training with Additional Self-Generated Code}
\label{sec:code_evol}
\paragraph{Performance cannot be steadily improved when the model is trained with both self-generated instructions and code.} 
We conduct the following two experiments to examine whether training with the code generated by the original model provides additional benefits.
\begin{enumerate}
\item \textbf{Code $\rightarrow$ NL $\rightarrow$ Code:} Regenerating new response code for the new instructions obtained by \xname. 
\item \textbf{Code $\rightarrow$ Code $\rightarrow$ NL:} Prompting the original model to give more complex code and applying \xname to the new code. 
\end{enumerate}
The results are shown in Table \ref{tab:code_evol}. Unstable performance reveals issues with the quality of the self-generated code of original models.

\begin{table}[t]
    \centering
    \setlength{\tabcolsep}{1mm}

        \begin{tabular}{lcc}
        \toprule
        Data-Generation Method & HumanEval (+) & MBPP (+)     \\
        \midrule
\rowcolor[rgb]{0.925,0.925,0.925} Code $\rightarrow$ NL (ours) & \textbf{76.2} (\textbf{72.0}) & \underline{70.6} (\underline{60.1}) \\
        Code $\rightarrow$ NL $\rightarrow$ Code  & \underline{73.2} (\underline{68.9}) & 67.7 (57.7) \\
        Code $\rightarrow$ Code $\rightarrow$ NL  & \underline{73.2} (68.3) & \textbf{70.9} (\textbf{62.2}) \\
        \bottomrule
        \end{tabular}
    
    \caption{\textbf{Comparison of \xname with other alternative data generation methods which prompt the original model to generate additional code (for CL-7B).}}
    \label{tab:code_evol}
\end{table}

\section{Conclusion}
\label{sec:conclusion}
In conclusion, this paper presents a novel approach to enhancing the capabilities of open-source code LLMs by leveraging self-generated data for instruction tuning, rather than relying solely on data from powerful closed-source LLMs like GPT-3.5 and GPT-4. Our proposed method, named \xname, capitalizes on the inherent asymmetry in translating between formal and informal languages. By reversing the conventional process, \xname generates additional natural language instructions from code snippets via summarization and self-evaluation techniques. The effectiveness of this methodology is demonstrated through the development of \modelname, a new series of code LLMs that not only outperform their predecessors in traditional benchmarks but also show significant improvement across diverse coding tasks.
\section*{Acknowledgements}
\label{sec:acknowledgements}

We thank Lei Qi for helping us analyze data and convert NL to code in sanity check experiments (Section \ref{sec:insights}) during the rebuttal.

This work is partially supported by the National Key R\&D Program of China (under Grant 2022YFB4501600), the NSF of China (under Grants 61925208, U22A2028, 6240073476, 62222214, 62341411, 62102398, 62102399, 62302478, 62302482, 62302483, 62302480,62302481), Strategic Priority Research Program of the Chinese Academy of Sciences, (Grant No.XDB0660200, XDB0660201, XDB0660202), CAS Project for Young Scientists in Basic Research (YSBR-029), Youth Innovation Promotion Association CAS and Xplore Prize.
\bibliography{aaai25}

\clearpage
\FloatBarrier
\appendix



\section{Limitations}
\label{sec:limitations}
Our research is subject to two primary limitations. Firstly, \xname may perform worse on models with weaker Code-to-NL ability because the improvement of our method stems from the advantage of Code-to-NL over NL-to-Code.
A promising avenue for future exploration is to investigate the disparities in the model's capabilities across different modalities (e.g., Code-to-Code, Code-to-NL, NL-to-Code), which may enable the development of more advanced language models that surpass current performance ceilings. 
Secondly, synthetic data generation relies on access to high-quality code snippet datasets. 
Future work should focus on reducing the required code snippet volume to enhance efficiency and feasibility.

\section{Method Details}
\label{app:method_details}

\paragraph{Algorithm Workflow.} The algorithm workflow of \xname is shown in Algorithm \ref{fig:algorithm_workflow} (see Page 15).

\paragraph{Prompts.} We show the prompts of \xname for code summarization, self-evaluation, and instruction-tuning in Figure~\ref{fig:prompt} (see Page 15).

\paragraph{Examples.} Figure~\ref{fig:multiple-code-response-example} and Figure~\ref{fig:bad-instruction-example} are examples for the details in \xname.
Figure~\ref{fig:multiple-code-response-example} is an example of a response with multiple code snippets and Figure~\ref{fig:bad-instruction-example} is an example of a summarization mistake.

\addtocounter{figure}{1}

\begin{figure}[t]
  \centering
\includegraphics[width=\linewidth]{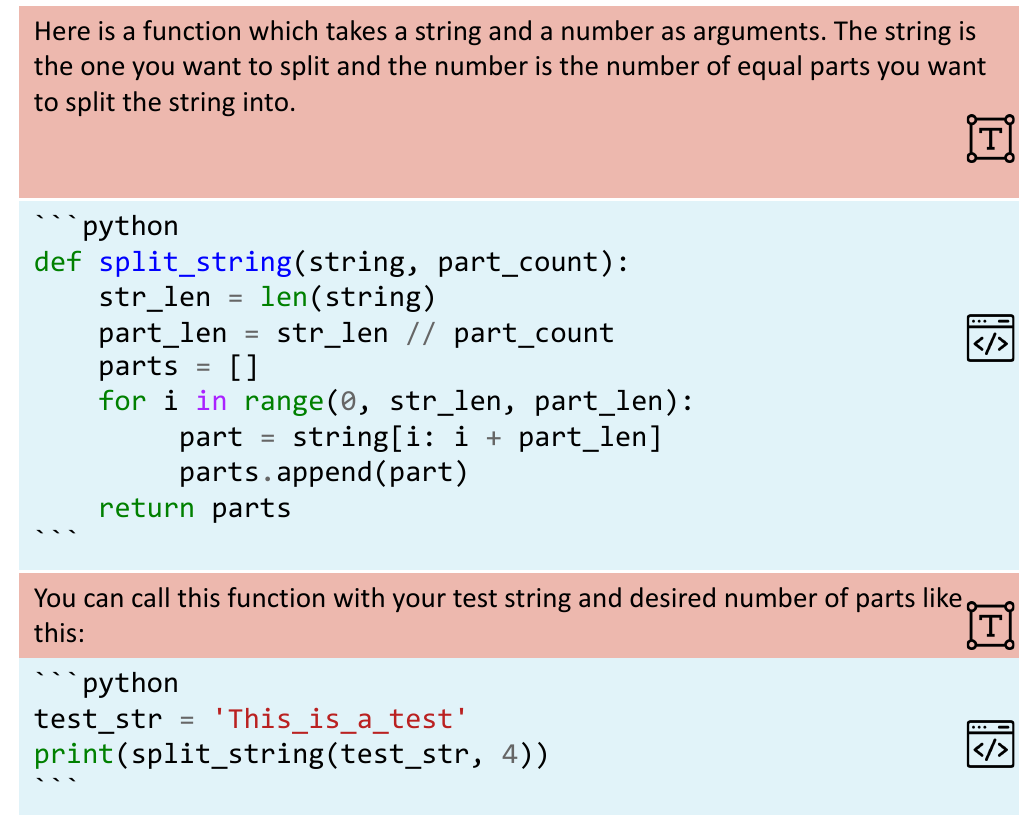}
\caption{An example response with multiple parts of code.}
\label{fig:multiple-code-response-example}
\end{figure}

\begin{figure}[t]
  \centering
\includegraphics[width=\linewidth]{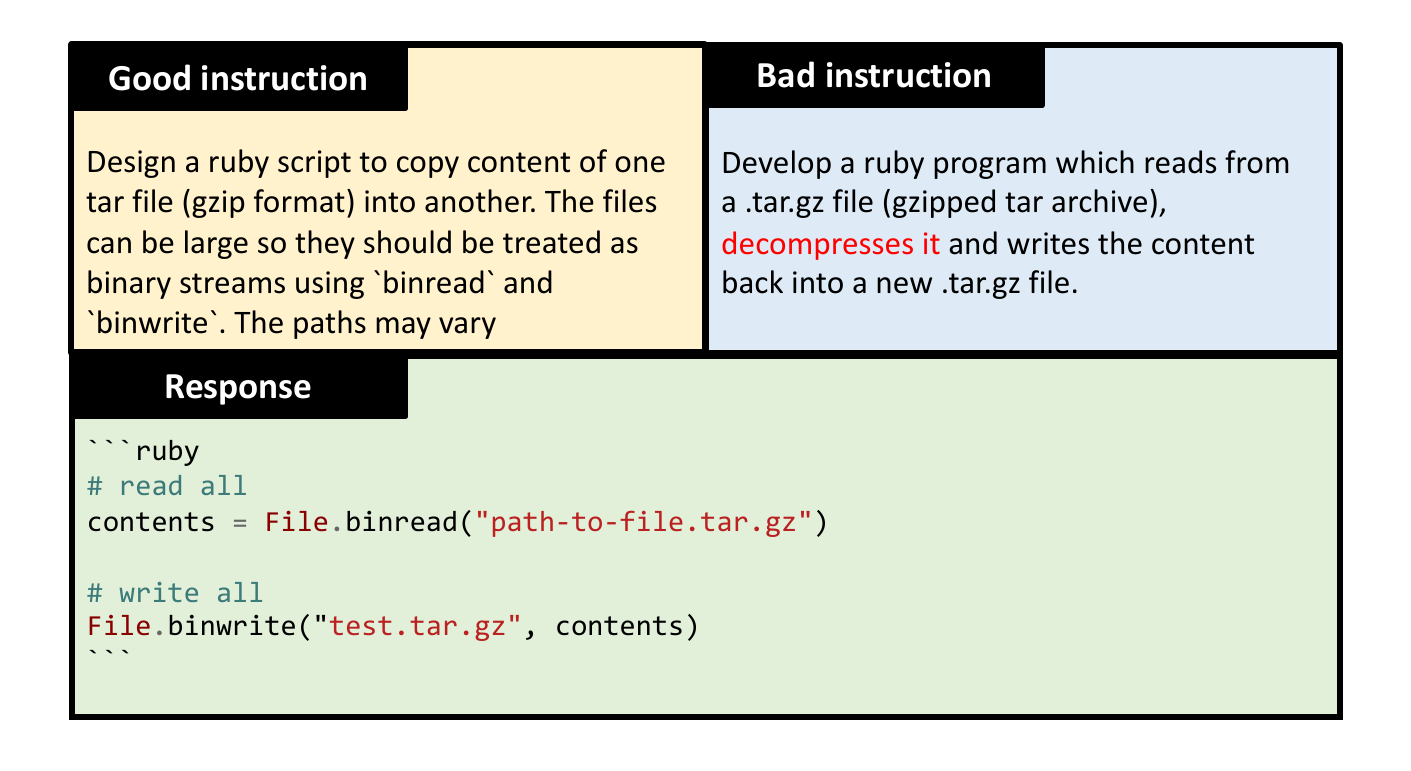}
\caption{An example of a summarization mistake.}
\label{fig:bad-instruction-example}
\end{figure}

\section{Further Analysis}
\label{app:further_analysis}

We conducted several further analyses of Inverse-Instruct including:
\begin{itemize}
    \item The dataset's statistical characteristics (Appendix \ref{app:dataset_analysis}).
    \item The way to quantify the mismatch between an LLM's code generation and code summarization ability (Appendix \ref{sec:self_consistency}).
    \item Self-improvement experiments using only base models (Appendix \ref{sec:self_improving}).
    \item The comparison between \xname and StarCoder2-Instruct (Section \ref{sec:compare_with_starcode2}).
    \item An explanation for the different training settings in baselines, and the performance of alternative training orders (Appendix \ref{sec:training_order}).
    \item Diversity analysis for instructions generated by \xname (Appendix \ref{sec:diversity_of_generated_instructions}).
\end{itemize}

\subsection{Dataset Analysis}
\label{app:dataset_analysis}

Following Magicoder \cite{magicoder}, we conduct further analysis for the datasets.

\paragraph{Categories of Instructions.} We use the text embeddings generated by \textsc{instructor} \cite{instructor} to analyze categories of the instructions generated by \xname. We calculate the ratios of 10 coding-related categories of the instructions in \evolcode and the dataset generated by WizardCoder-GPT4-CL-7B. The results are illustrated in Figure~\ref{fig:evol-gpt4-categories} and Figure~\ref{fig:codellama-categories} (see Page 16).

\paragraph{Length Distribution of the Datasets.} We depict the length distribution by counting the token for the instructions and responses in \evolcode and the dataset with instructions generated by WizardCoder-GPT4-CL-7B. The distributions are shown in Figure~\ref{fig:evol-length} and Figure~\ref{fig:codellama-length}. It is noticed that the responses in the new dataset are much shorter since they are pure code snippets extracted from the original dataset. 

\addtocounter{figure}{2}

\begin{figure}[t] 
    \centering
    \includegraphics[width=0.8\linewidth]{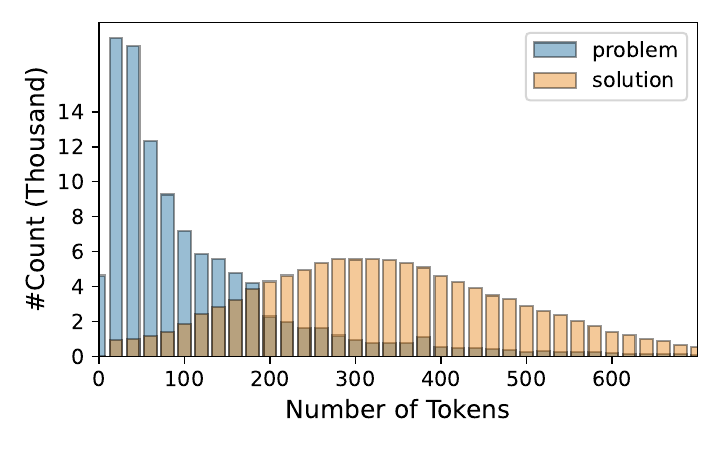}
    \caption{Token count distribution of the instructions and responses in original dataset \evolcode.}
    \label{fig:evol-length}
\end{figure}

\begin{figure}[t]
    \centering
    \includegraphics[width=0.8\linewidth]{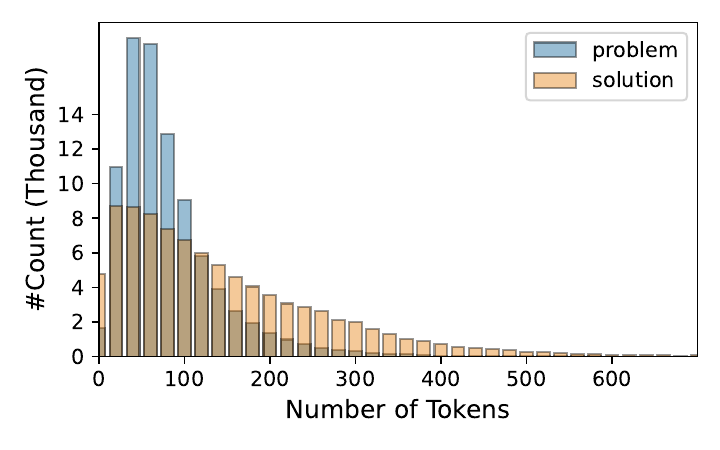}
    \caption{Token count distribution of the instructions and responses in the dataset generated by WizardCoder-GPT4-CL-7B.}
    \label{fig:codellama-length}
\end{figure}

\begin{figure}[t]
  \centering
\includegraphics[width=0.8\linewidth]{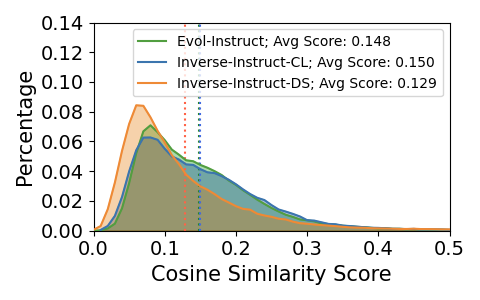}
\caption{Cosine similarities between HumanEval and different datasets. ``Evol-Instruct'' represents the original dataset \evolcode.}
\label{fig:similarity}
\end{figure}

\paragraph{Similarity with HumanEval.} We compute the cosine similarity between the HumanEval benchmark and our datasets using TF-IDF \cite{tfidf} embeddings. The results are shown in Figure~\ref{fig:similarity}, which demonstrates that the datasets have a low similarity to the benchmark.

\subsection{Self-Consistency on Generation and Summarization}
\label{sec:self_consistency}

\paragraph{We attempt to quantify the mismatch between the LLM's code generation and code summarization ability by measuring \textit{self-consistency}.}
That is, whether the model can generate equivalent code after summarizing the code it generates before.
Self-consistency is defined as 
\begin{equation*}
    \indicator({M(M^{-1}(M(x)))=M(x)}),
\end{equation*}
where $M(\cdot)$ denotes the LLM's code generation process, $M^{-1}(\cdot)$ denotes the LLM's code summarization process, $\indicator(\cdot)$ is the indicator function, and $=$ denotes functional equivalence.

Specifically, we use the benchmark MBPP+ to measure the self-consistency  in three steps:
Firstly, we prompt the code LLM $M$ with problems $x_i$ to generate code $y_i=M(x_i)$.
Then, code LLM summarizes the code $y_i$ to get new instructions $x'_i=M^{-1}(y_i)$. 
Next, we let the code LLM generate new code $y'_i$ based on $x'_i$, \ie $y'_i=M(x'_i)$. Finally, we evaluate if $y'_i$ and $y_i$ are functional equivalents by measuring their outputs given the same inputs taken from the benchmark. 
The result is calculated as the pass@1 of new code responses $y'_i$ taking the original code responses $y_i$ as ground truth solutions.

The results are shown in Table \ref{tab:self-consistency}. 
\begin{table}[h]
\centering
\begin{tabular}{lc}
\toprule
Model                  & Self-Consistency \\
\midrule
WizardCoder-GPT4-CL    & \underline{69.0} (\underline{65.6})       \\
 \rowcolor[rgb]{0.925,0.925,0.925}InverseCoder-CL (ours) & \textbf{76.0 (73.5)}     \\
\midrule
WizardCoder-GPT4-DS    & \underline{76.1} (\underline{69.7})      \\
 \rowcolor[rgb]{0.925,0.925,0.925}InverseCoder-DS (ours) & \textbf{80.4 (75.3)}             \\
\bottomrule
\end{tabular}
 \caption{\Passat{1} (\%) results on MBPP(+) in the self-consistency experiment. We remove the problems whose inputs will
 cause a runtime error when applied to the original code response $y_i$.}
\label{tab:self-consistency}
\end{table}
\modelname has better self-consistency than the original models, which indicates that the original models have a larger gap between generation and summarization, and the performance improvement of \modelname may come from bridging this gap. 

\subsection{Comparison with StarCoder2-Instruct}
\label{sec:compare_with_starcode2}

\paragraph{\xname surpasses another self-improvement method: StarCoder2-Instruct \cite{StarCoder2-Instruct}.} Most recently, StarCoder2-Instruct has been introduced as a self-improvement method specifically designed for code LLMs. It uses a similar pipeline to Magicoder \cite{magicoder} but with the additional step of generating concepts from code before creating problems. The detailed comparison between Inverse-Instruct and StarCoder2-Instruct is shown in Table \ref{tab:comparsion_with_starcoder2}. We find that \xname takes less time and produces a better instruction-tuned model.

\begin{table}[h]
\centering
\setlength{\tabcolsep}{1mm}
\fontsize{9}{11}\selectfont
\begin{tabular}{lccc}
\toprule
Model & HumanEval (+) & MBPP(+) & Time Cost (h) \\
\midrule
Starcoder2-Instruct & \underline{71.3} (\underline{66.5})                     & \underline{69.6} (\underline{59.8}) &  \underline{80}              \\
 \rowcolor[rgb]{0.925,0.925,0.925}InverseCoder (ours)        & \textbf{76.2 (72.0)}                     & \textbf{70.6 (60.1)} & \textbf{4}              \\
\bottomrule
\end{tabular}
\caption{\textbf{The comparison between our method with Starcoder2-Instruct.} We reproduce it using the same base model (CodeLlama-Python-7B), dataset (we use the response code as seed) and training settings as ours.}
\label{tab:comparsion_with_starcoder2}
\end{table}

\subsection{Self-improving for Base Models}
\label{sec:self_improving}
\paragraph{Base models can improve themselves by generating instructions from source code through \xname.} We replace the instruction-tuned code LLM with its base model in our pipeline to demonstrate that a base model can self-improve solely by relying on source code. Specifically, we use the preprocessed code responses $\{y_i^*\}$ as the unlabeled source code. Then, we apply code summarization and self-evaluation to it using a base model. Finally, we fine-tune the base model only with the instruction data generated by itself. The enhanced performance of the base model (Table \ref{tab:self-improve}) reveals that it is useful to apply \xname for the base model in a situation with adequate unlabeled data but insufficient human-annotated instructions.

\begin{table}[h]
\centering
\begin{tabular}{ccc}
\toprule
Model                       & HumanEval (+) & MBPP (+)     \\
\midrule
CodeLlama-Python-7B         & 39.6 (35.4)  & 51.9 (43.7) \\
+$1$ epoch self-generated & \underline{54.3} (\underline{49.3})  & \underline{52.9} (\underline{44.7}) \\
+$2$ epoch self-generated & \textbf{54.9 (50.6)}  & \textbf{54.2 (46.6)} \\
\bottomrule
\end{tabular}
 \caption{\Passat{1} (\%) results on HumanEval~(+) and MBPP~(+) in self-improving  experiments for base model \codellamapy-7B. We report performances of the model finetuned with generated data for $1$ and $2$ epochs.}
 \label{tab:self-improve}
\end{table}

\subsection{Explanation for Different Training Settings Between \modelname and Baselines}
\label{sec:training_order}

\paragraph{It is fair to compare \modelname with baselines in different training settings.}
Since the different training settings (epochs, orders for datasets) are also used in previous works (see Table \ref{tab:training_settings}), we follow them to report the best results among different training settings in our paper. To validate this fact, we also train \magicoders-CL in our epoch setting (only train 1 epoch for the lower-quality dataset Magicoder-OSS-Instruct-75K). The results (Table \ref{tab:magicoder_different}) show that the model's performance dropped in our setting.
\begin{table}[h]
\centering
\begin{tabular}{ll}
\toprule
Model               & Training Settings               \\
\midrule
WizardCoder         & 200 steps for all data          \\
MagicoderS          & 2 epochs weak + 2 epochs strong \\
WaveCoder-Ultra     & 3 epochs for all data           \\
OpenCodeInterpreter & 3 epochs for all data           \\
\rowcolor[rgb]{0.925,0.925,0.925} InverseCoder (ours) & 1 epoch weak + 2 epochs strong  \\
\bottomrule
\end{tabular}
\caption{\textbf{Training settings of different instruction-tuned code LLMs.}}
\label{tab:training_settings}
\end{table}

\begin{table}[h]
\centering
\setlength{\tabcolsep}{1mm}
\fontsize{9}{11}\selectfont
\begin{tabular}{lc}
\toprule
Method                                    & HumanEval (+) \\
\midrule
2 epochs weak + 2 epochs strong (original) & 70.7 (66.5)   \\
1 epochs weak + 2 epochs strong           & 68.3 (64.6)   \\
\bottomrule
\end{tabular}
\caption{\textbf{Performance degradation after training \magicoders-CL with our training epoch setting.}}
\label{tab:magicoder_different}
\end{table}

\paragraph{We explore different settings and find that the strong-after-weak order performs the best among three settings.} 
We use the training order of \magicoders \cite{magicoder}, which starts with data generated by a weaker model and then uses data generated by a stronger model. We show the performance results of other training orders in Table \ref{tab:training_order} to demonstrate our training order's effectiveness.  
\begin{table}[h]
\centering
\setlength{\tabcolsep}{0.5mm}
\fontsize{9}{11}\selectfont
\begin{tabular}{lc}
\toprule
Training Order                          & HumanEval (+) \\
\midrule
GPT-4 data (baseline) & \underline{72.6} (\underline{68.9}) \\
The mixture of GPT-4 and self-generated data  & \underline{72.6} (64.6)   \\
GPT-4 data + self-generated data        & 64.0 (59.1)   \\
\rowcolor[rgb]{0.925,0.925,0.925} Self-generated data + GPT-4 data (ours) & \textbf{76.2 (72.0)}  \\
\bottomrule
\end{tabular}
 \caption{\textbf{The comparison between our training order and other alternatives.} We use CodeLlama-Python-7B as the base model and train for a total of 3 epochs under each order.}
 \label{tab:training_order}
\end{table}

\subsection{Diversity Analysis for Generated Instructions}
\label{sec:diversity_of_generated_instructions}
To compare the diversity of the data generated by Code-to-NL with that generated by NL-to-Code, we reproduce the Diverse Instruction Tuning (DIT) method of DolphCoder \cite{wang-etal-2024-dolphcoder}, which uses different system prompts to obtain diverse code responses. The results (Table \ref{tab:comparsion_with_dolphcoder}) show that the performance of DIT is worse than \xname. We believe that it is because the Code-to-NL method generates more diverse data, and we support this idea in two aspects:
\begin{table}[h]
\centering
\begin{tabular}{ccc}
\toprule
Method                 & HumanEval (+) & MBPP (+)    \\
\midrule
DIT                    & 69.5 (66.5)   & 69.8 (58.7) \\
\rowcolor[rgb]{0.925,0.925,0.925} Inverse-Instruct (ours) & \textbf{76.2 (72.0)}  & \textbf{70.6 (60.1)} \\
\bottomrule
\end{tabular}
\caption{\textbf{The comparison between \xname and Diverse Instruction Tuning (DIT).} We generate the new code by DIT using WizardCoder-GPT4-CL and \evolcode, then apply data selection to it. Other settings are the same as \xname.}
\label{tab:comparsion_with_dolphcoder}
\end{table}

\begin{figure}[h]
    \centering
    \includegraphics[width=\linewidth]{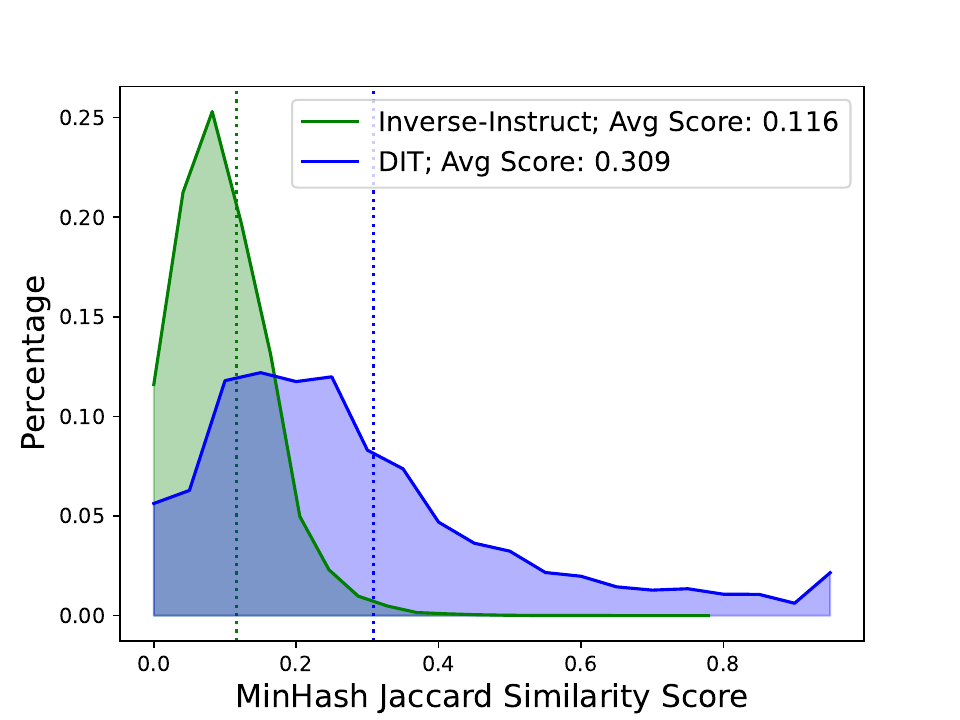}
    \caption{\textbf{\xname vs. DIT: MinHash Jaccard similarity between generated data and the original one.}}
    \label{fig:minhash_sim}
\end{figure}

\begin{figure}[h]
    \centering
    \includegraphics[width=\linewidth]{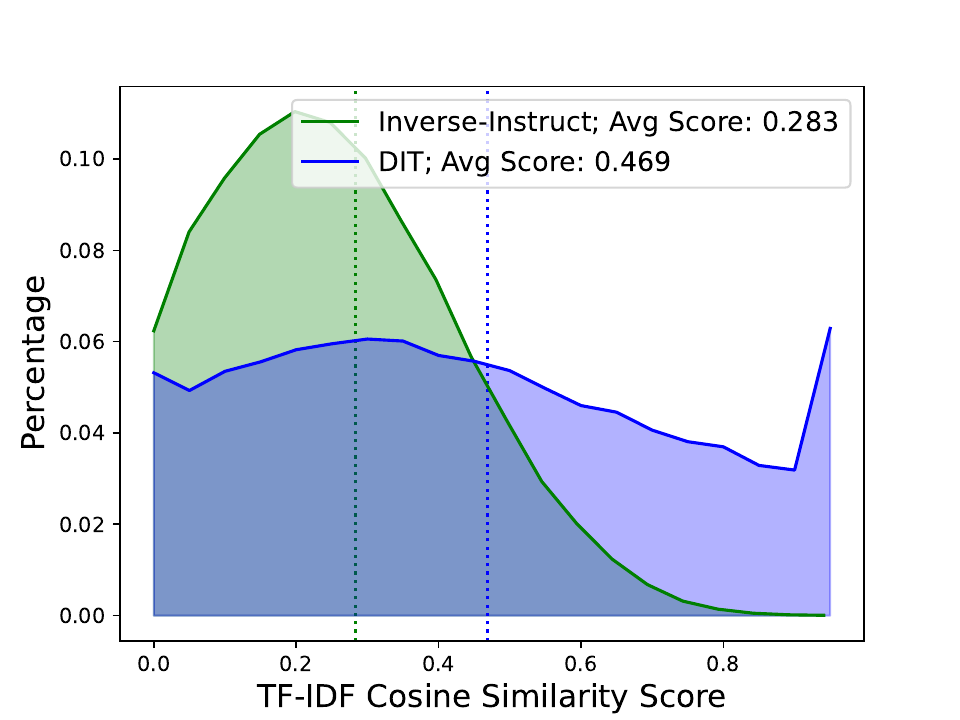}
    \caption{\textbf{\xname vs. DIT: TF-IDF cosine similarity between generated data and the original one.}}
    \label{fig:tfidf_sim}
\end{figure}

\paragraph{The instructions generated by \xname (Code-to-NL) are more diverse than the code generated by DIT (NL-to-Code).}
We calculate the Jaccard similarity of MinHash \cite{minhash} and cosine similarity of TF-IDF embeddings \cite{tfidf} between each generated instruction and its original one for \xname, and the same similarities between each generated response and its original one for DIT. The results are shown in Figures \ref{fig:minhash_sim} and \ref{fig:tfidf_sim}, which demonstrates that the generated instructions are less similar to the original ones compared to the generated code.

\begin{table}[h]
\centering
\fontsize{9}{11}\selectfont
\begin{tabular}{cccc}
\toprule
\multicolumn{1}{c}{\multirow{2}{*}{Method}} & \multicolumn{3}{c}{Threshold} \\\cmidrule(r){2-4}
                                            & 0.6      & 0.7      & 0.8     \\
\midrule
DIT                                         & 8.75   & 3.53   & 1.48  \\
\rowcolor[rgb]{0.925,0.925,0.925} Inverse-Instruct (ours)                     & \textbf{7.17}   & \textbf{1.40}   & \textbf{0.38}  \\
\bottomrule
\end{tabular}
\caption{\textbf{The overall data duplication rate (\%) of \xname and DIT.} We use three different similarity thresholds for MinHash LSH keeping $\texttt{num\_perm}=128$.}
\label{tab:minhash_lsh}
\end{table}
\paragraph{\xname tends to produce the datasets with less redundancy.}
We use MinHash LSH \cite{datasketch} to deduplicate the entire dataset (110K GPT-4 instruction-response pairs and 90K self-generated pairs) for both DIT and \xname. Table \ref{tab:minhash_lsh} is the duplication rate under different thresholds, which shows that the entire dataset of \xname is more diverse.

\section{Generation Examples}
\label{app:generation_examples}
Tables \ref{tab:example1} and \ref{tab:example2} (Page 17 and 18) are examples of the responses given by \modelname. 

\newcommand{\InlineComment}[1]{\(\triangleleft\) #1}

\begin{figure*}[h!]
\centering
\begin{minipage}{0.75\textwidth}

\begin{algorithm}[H]
\caption{Algorithm Workflow of \xname}
\label{fig:algorithm_workflow}
\textbf{Input}: An instruction-tuned code LLM $M$ and its instruction-tuning dataset $\mathcal{D}$. \\
\textbf{Parameter}: The number of summarization samples $K$. An instruction prefix pool $P$. \\
\textbf{Output}: An augmented dataset $\mathcal{D}^*$.
\begin{algorithmic}[1] 

\State $\mathcal{C} \gets \emptyset$. \Comment{Initialize the code corpus}
\For{each instruction-response pair $(x_i, y_i)$ in $\mathcal{D}$} \Comment{Preprocess for $\mathcal{D}$}
    \If {there are code snippets in $y_i$}
        \State $y^*_i \gets$ the first part of the code in $y_i$.
        \State $\mathcal{C} \gets \mathcal{C} \cup \{y^*_i\}$.
    \EndIf
\EndFor

\State $\mathcal{D}^* \gets \emptyset$. \Comment{Initialize the augmented dataset}
\For{each code snippet $y^*_i$ in $\mathcal{C}$}
    \State $\mathcal{I}_i^* \gets \emptyset$
    \Comment{Initialize the instruction set for $y^*_i$}
    \For{$j \gets 1$ to $K$}
        \State $p_j \gets$ $\textit{random\_choice}(P)$ \Comment{Randomly select a prefix from the pool $P$}
        \State $x^*_{ij} \gets$ $M(y^*_i, p_j)$. \Comment{Generate a new instruction starting with $p_j$ for $y^*_i$}
        \State $s^*_{ij} \gets$ $M(x^*_{ij}, y_i^*)$. \Comment{Evaluate the correctness score of $y_i^*$ under $x^*_{ij}$}
        \State $\mathcal{I}_i^* \gets \mathcal{I}_i^* \cup \{(x^*_{ij}, s^*_{ij})\}$. 
    \EndFor
    \State $(x_i^{**}, s_{i}^{**}) \gets (x^*_{ij}, s^*_{ij})$ where $s^*_{ij} = \max_{\mathcal{I}_i^*} s^*_{ij}$. \Comment{Select the best instruction in $\mathcal{I}_i^*$}
    \State $\mathcal{D}^* \gets \mathcal{D}^* \cup \{(x_i^{**}, y_i^*)\}$.
\EndFor
\State \Return $\mathcal{D}^*$


\end{algorithmic}
\end{algorithm}

\end{minipage}
\end{figure*}

\addtocounter{figure}{-10}
\begin{figure*}[t]
  \centering
\includegraphics[width=0.97\linewidth]{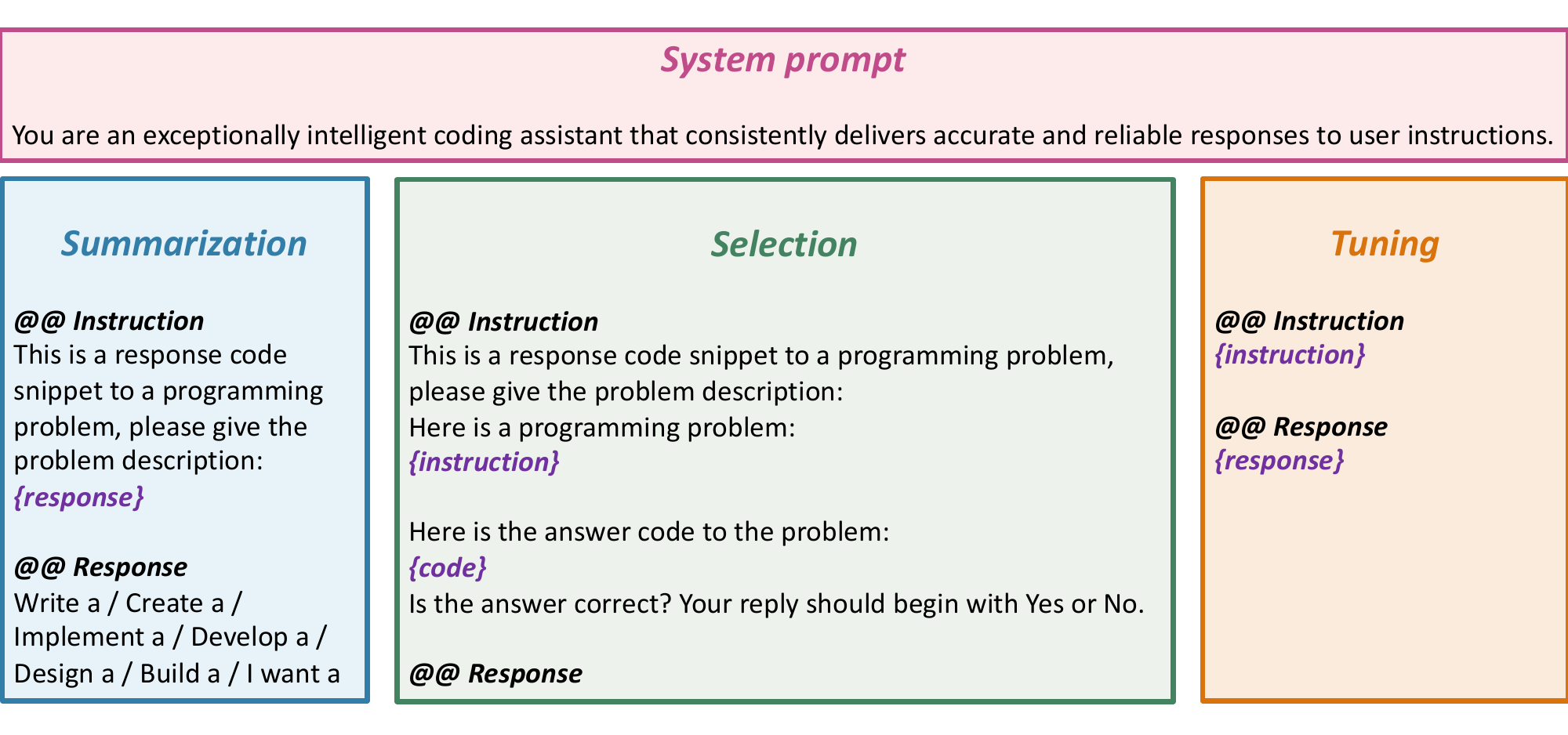}
\caption{\textbf{The prompts of \xname for code summarization, self-evaluation, and instruction-tuning.} For code summarization, we use a diverse set of initial verbs in the prefixes to ensure the overall diversity of the instructions. We first count the first verb frequencies of each instruction in the original dataset and choose the top 5 most frequent verbs. Then we ask ChatGPT to give similar verbs to expand the first verb pool for prompt prefixes.
}
\label{fig:prompt}
\end{figure*}
\addtocounter{figure}{10}

\addtocounter{figure}{-8}
\begin{figure*}[t]
  \centering
\includegraphics[width=1\linewidth]{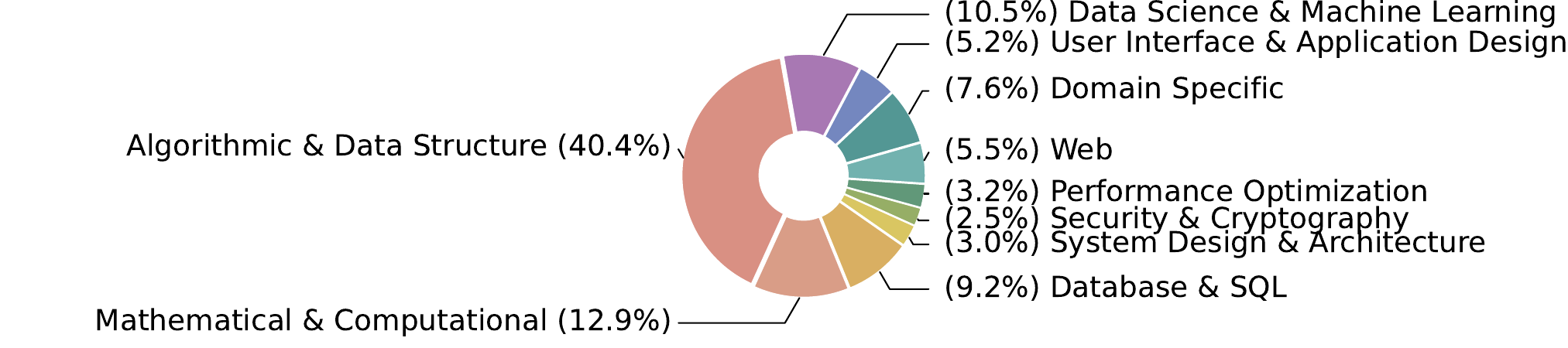}
\caption{Instruction category constitution of \evolcode.}
\label{fig:evol-gpt4-categories}
\end{figure*}

\begin{figure*}[t]
  \centering
\includegraphics[width=\linewidth]{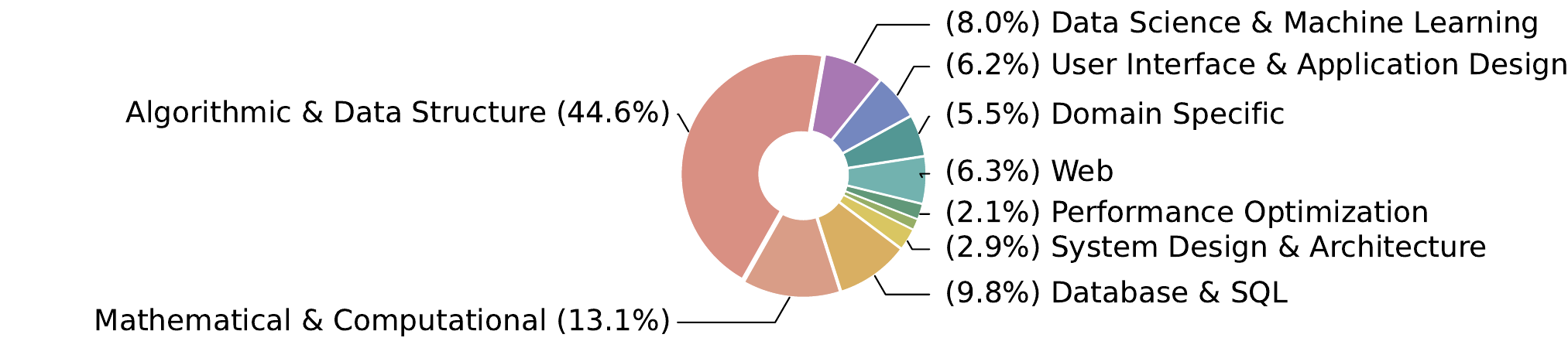}
\caption{Instruction category constitution of the dataset generated by WizardCoder-GPT4-CL-7B.}
\label{fig:codellama-categories}
\end{figure*}
\addtocounter{figure}{8}

\begin{table*}[!htb]
    \centering
    \begin{tabular}{p{2.5cm}|p{10cm}}
        \toprule
        \textbf{Instruction} & \textbf{Response}\\
        \midrule
        Write a Python script that lists all the files in the current directory. &
        \begin{minipage}{29em}
            Here is a simple Python script that lists all the files in the current directory:

\lstset{style=codestyle}
\begin{lstlisting}[language=Python]
import os

# Get the current directory
current_dir = os.getcwd()

# List all files in the current directory
files = os.listdir(current_dir)

# Print the files
for file in files:
    print(file)
\end{lstlisting}

This script uses the \`{}os\`{} module in Python, which provides a way of using operating system dependent functionality. The \`{}os.getcwd()\`{} function returns the current working directory, and \`{}os.listdir(path)\`{} returns a list containing the names of the entries in the directory given by path.                              
        \end{minipage}\\
        \midrule
        Please show me how to calculate the first ten primes in cpp. &
        \begin{minipage}{29em}
Here is a simple C++ program that calculates the first ten prime numbers:
\lstset{style=codestyle}
\begin{lstlisting}[language=C++]
#include<iostream>
using namespace std;

bool isPrime(int num) {
    if (num <= 1)
        return false;
    for (int i = 2; i * i <= num; i++) {
        if (num % i == 0)
            return false;
    }
    return true;
}
int main() {
    int count = 0;
    int num = 2;
    while(count < 10) {
        if(isPrime(num)) {
            cout << num << " ";
            count++;
        }
        num++;
    }
    return 0;
}
\end{lstlisting}

This program defines a function \`{}isPrime\`{} that checks if a number is prime. It then uses this function in a loop in the \`{}main\`{} function to find and print the first ten prime numbers.

        \end{minipage}\\
        \bottomrule
    \end{tabular}
    \caption{Instruction and response examples of \modelname{}.}
    \label{tab:example1}
\end{table*}
\begin{table*}[!htb]
    \centering
    \begin{tabular}{p{2.5cm}|p{10cm}}
        \toprule
        \textbf{Instruction} & \textbf{Response}\\
        \midrule
        How to do matrix multiplication in Pytorch? Show me an example. &
        \begin{minipage}{29em}
            Matrix multiplication in PyTorch can be done using the \`{}mm\`{} function from the \`{}torch\`{} module. Here is an example:

\lstset{style=codestyle}
\begin{lstlisting}[language=Python]
import torch

# Create two matrices
matrix1 = torch.tensor([[1, 2, 3], [4, 5, 6]])
matrix2 = torch.tensor([[7, 8], [9, 10], [11, 12]])

# Perform matrix multiplication
result = torch.mm(matrix1, matrix2)

print(result)
\end{lstlisting}

In this example, \`{}matrix1\`{} is a 2x3 matrix and \`{}matrix2\`{} is a 3x2 matrix. The \`{}mm\`{} function performs matrix multiplication on these two matrices and returns a new matrix.

The output will be:

\lstset{style=codestyle}
\begin{lstlisting}[language=Python]
tensor([[ 58,  64],
        [139, 154]])
\end{lstlisting}

        \end{minipage}\\
        \midrule
        Please check if the network can connect to www.example.com using Python. &
        \begin{minipage}{29em}

Sure, you can use the \`{}requests\`{} library in Python to check if a network can connect to \`{}www.example.com\`{}. Here is a simple script:
\lstset{style=codestyle}
\begin{lstlisting}[language=Python]
import requests

try:
    response = requests.get("http://www.example.com")
    response.raise_for_status()
    print("Connection successful")
except requests.exceptions.HTTPError as errh:
    print ("Http Error:",errh)
except requests.exceptions.ConnectionError as errc:
    print ("Error Connecting:",errc)
except requests.exceptions.Timeout as errt:
    print ("Timeout Error:",errt)
except requests.exceptions.RequestException as err:
    print ("Something went wrong",err)
\end{lstlisting}
This script sends a GET request to \`{}www.example.com\`{} and checks for any exceptions that might occur during the request. If the request is successful, it prints ``Connection successful''. If there is an HTTP error, a connection error, a timeout error, or any other type of request exception, it prints the error.

        \end{minipage}\\
        \bottomrule
    \end{tabular}
    \caption{Instruction and response examples of \modelname{}.}
    \label{tab:example2}
\end{table*}

\end{document}